\documentclass[3p,times]{elsarticle}

\usepackage{ecrc}
\usepackage{array}
\usepackage{algorithmic}
\usepackage{graphicx}
\usepackage{textcomp}
\usepackage{xcolor}
\usepackage{subcaption}
\usepackage{natbib}

\newcolumntype{P}[1]{>{\centering\arraybackslash}p{#1}}

\def\BibTeX{{\rm B\kern-.05em{\sc i\kern-.025em b}\kern-.08em
T\kern-.1667em\lower.7ex\hbox{E}\kern-.125emX}}

\volume{00}

\firstpage{1}

\runauth{}

\usepackage{amssymb}

\usepackage[figuresright]{rotating}

\begin{document}

\begin{frontmatter}




\title{The Penetration of Internet of Things in Robotics: Towards a Web of Robotic Things}

 \author[label1,label2]{Andreas Kamilaris}
  \author[label3]{Nicol\`{o} Botteghi}
 \address[label1]{Research Centre on Interactive Media, Smart Systems and Emerging Technologies (RISE), Nicosia, Cyprus}
  \address[label2]{Dept. of Computer Science, University of Twente, Enschede, The Netherlands}
  \address[label3]{Dept. of Electrical Engineering, University of Twente, Enschede, The Netherlands}

\begin{abstract}
As the Internet of Things (IoT) penetrates different domains and application areas, it has recently entered also the world of robotics.
Robotics constitutes a modern and fast-evolving technology, increasingly being used in industrial, commercial and domestic settings.
IoT, together with the Web of Things (WoT) could provide many benefits to robotic systems. Some of the benefits of IoT in robotics have been discussed in related work. This paper moves one step further, studying the actual current use of IoT in robotics, through various real-world examples encountered through a bibliographic research. The paper also examines the potential of WoT, together with robotic systems, investigating which concepts, characteristics, architectures, hardware, software and communication methods of IoT are used in existing robotic systems, which sensors and actions are incorporated in IoT-based robots, as well as in which application areas. Finally, the current application of WoT in robotics is examined and discussed.
\end{abstract}


\end{frontmatter}

\newpage

\section{Introduction}
The Internet of things (IoT) \cite{weber2010internet} is the extension of Internet connectivity into physical devices and everyday objects. 
As the IoT penetrates different domains, application areas and scientific disciplines \cite{miorandi2012internet},
it is worth the effort to examine its impact, interaction and application in the research area of robotics.

\begin{figure}[h]
   \centering
   \includegraphics[width=0.40\columnwidth]{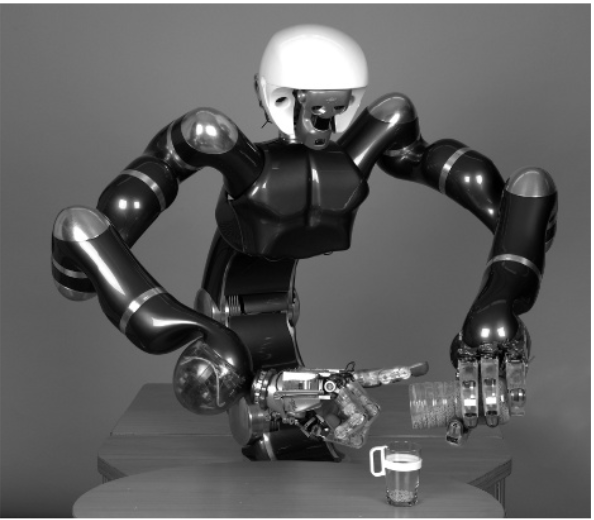}
   \caption{An example robot pouring some liquid from one cup to another (Source: \cite{siciliano2010robotics}).}
   \label{fig:robotex1}
   \vspace{-0.3cm}
\end{figure}

Robotics constitutes a modern and fast-evolving technology \cite{siciliano2010robotics}, which is increasingly being used in industrial, commercial and domestic settings, as well as for rescue operations where there are safety risks for humans \cite{vertut2013teleoperation}. Robotics can be defined as the branch of engineering that involves the conception, design, manufacture and operation of robots \cite{fuller1998robotics}. \textit{Robot} comes from the Czech word \textit{robota} which means forced work or labour. The word robot today means any man-made machine that can perform work or other actions normally performed by humans, either automatically or by remote control. Robots are employed because it is often cheaper to use them over humans, easier for robots to do some job and sometimes the only possible way to accomplish some tasks. Most robots are composed of a controller (i.e. the \textit{brain} of the robot), mechanics (i.e. motors, pistons, grippers, wheels and gears that make the robot move, grab, turn, and lift) and sensors, i.e. to help the robot perceive its surroundings. An example of a robot with a human-like shape is provided in Figure \ref{fig:robotex1}. Although robots have been mostly used in industrial applications till date, recent technological progress in the emerging domains of cognition, manipulation and interactions is moving the robotics industry toward service robots and human-centric design \cite{pwcReport}.

The interconnection and relationship between the IoT and robotics has been defined as \textit{Internet of Robotic Things} (IoRT) \cite{ray2016internet, vermesan2017internet, simoens2018internet}. IoRT is about \textit{a global infrastructure enabling advanced robotic services by interconnecting robotic things based on, existing and evolving, interoperable information and communication technologies such as cloud computing, cloud storage and other existing Internet technologies}. IoT allows robots to communicate by means of the IP protocol, especially its IPv6 version, which is designed for billions of Internet-connected objects \cite{jara2012glowbal}. Internet connection permits updating information (and possibly the robot's firmware) in real-time \cite{turcu2012integrating}, storing/processing data on the cloud and taking advantage of Internet protocols for security, authentication, data integrity, message routing etc. \cite{stallings2004computer, suo2012security}.

We argue that IoRT should also embrace the Web of Things (WoT) \cite{wilde2007putting, guinard2011internet}, which is about approaches, software architectural styles and programming patterns that allow real-world objects to be part of the World Wide Web. WoT could offer additional benefits to the IoT-enabling of robotic-based systems, especially in terms of higher interoperability among the robot's sensory components, but also \textit{robot-to-robot} (R2R) and \textit{robot-to-human} (R2H) communication at the application layer. Additionally, WoT could facilitate the combination of Web services and robotic services, towards physical mashups \cite{guinard2009towards}, which could evolve to \textit{robotic mashups}, realizing the notion of a \textit{Web of Robotic Things} (WoRT). These benefits are discussed in Section \ref{WoT}, in more detail, together with a small historical journey on the first robots that had presence and allowed remote control via the Web, back in 1995.

The main motivation for this paper comes from the observation on how other research areas and/or application domains have been influenced in the past by the introduction of IoT and WoT. These areas include home automation \cite{stojkoska2017review, kamilaris2011homeweb}, smart buildings \cite{casini2014internet}, smart grid \cite{collier2016emerging}, remote healthcare \cite{yuehong2016internet}, agriculture ,\cite{Kamilaris2016wotplatforms} as well as a limited number of smart city applications \cite{Kamilaris2017a, KamilarisIeeeIoT16}. Examples of this influence by IoT/WoT involve reduced risks of vendor lock-in, adopting machinery (i.e. for agriculture) and sensing/automation systems from different companies, as these could become easily interoperable in the overall smart systems, easier data exchange among different, heterogeneous components, increased automation with less effort by means of Internet and Web standards etc. Use of open standards brings seamless connectivity and advanced interoperability, while the publishing of data produced by IoT sensors as open data on the Web promotes knowledge sharing and is important for the advancement of research in these fields.

Additional motivation for preparing this paper stems from the fact that related work in this field \cite{vermesan2017internet, simoens2018internet, ray2016internet, roy2017iot, grieco2014iot} has discussed only generally some of the opportunities of IoT in robotics. There is a gap in literature in relation to the actual degree of penetration of IoT in robotic systems and services, as well as how IoT has been used in robot systems till date. In addition, the current application and future potential of WoT in robotics has not been discussed in any relevant paper, except from \cite{grieco2014iot}, where the aspect of semantic consensus has been generally discussed, suggesting that this issue could be approached via Semantic Web technologies.
In particular, the work in \cite{simoens2018internet} focuses on the lower and higher level abilities of IoT-enabled robots, while Vermesan et al. \cite{vermesan2017internet} discuss generally some technologies of IoT that can support robotic systems. 
Ray \cite{ray2016internet} lists some examples of existing robots envisaged for an IoRT architecture, but the linking between the examples mentioned in the paper with the IoT is unclear and not convincing. The differences between this survey and relevant ones are summarized in Table \ref{tbl:comparison}.

\begin{table*}[htbp]
\caption{Diffferences between this survey and related ones.}
\begin{center}
\centering
\begin{tabular}{|p{4.0cm} | P{1.5cm} | P{2.0cm} | P{2.0cm} |  P{1.5cm} | P{2.7cm} |  P{1.3cm} |  }
\hline
\bf{Characteristic/Paper}  &  \bf{\cite{ray2016internet}} &  \bf{\cite{vermesan2017internet}} & \bf{ \cite{simoens2018internet}} &  \bf{\cite{roy2017iot}} &  \bf{\cite{grieco2014iot}} & \bf{Our paper}  \\
\hline
Review of papers demonstrating robots &  IoRT-based ones & & Only general & Only general & Only general & IoRT-based ones \\
\hline
Description of IoRT Technologies & X & Focus on software platforms and interoperability & & Only cloud robotics & IoT security, hardware platforms & X \\
\hline
Linking between robots, sensors, actions and applications &  & & Crossover of IoRT into nine robotic system abilities & Link with applications & Only general  & X \\
\hline
Linking of robots with IoRT technologies &  & & & & Link between IoRT technologies and EU projects & X  \\
\hline
Research challenges & X & & & & Only general  & X \\
\hline
Presenting the whole picture around IoRT & X & X & & & & X \\
\hline
Linking of robots with WoRT technologies &  & & & & Semantic consensus towards Semantic WoT & X \\
\hline
\end{tabular}
\label{tbl:comparison}
\end{center}
\end{table*}

Therefore, the contribution of this paper is to study the current use of IoT in robotics, through various real-world examples encountered through a bibliography-based research and to examine the possibilities of the WoT together with robotic systems. To the authors' knowledge, it is the most complete survey paper to date, aiming to present the actual research taking place in the intersection of these research domains.

The rest of this paper is organized as follows: Section \ref{meth} describes the methodology adopted
and Section \ref{robots} introduces the concept of robotics, clarifying its relationship with IoT.
Section \ref{IoT} presents the existing penetration of IoT in robotics, while Section \ref{WoT} studies the connection between WoT and robotic systems.
Finally, Section \ref{Disc} discusses the overall findings and Section \ref{conclusion} concludes the paper.

\section{Methodology}
\label{meth}
This paper aims to fill the aforementioned gaps in literature, by addressing the following research questions:
\begin{enumerate}
 \item Which concepts, characteristics, architectures, platforms, software, hardware and communication standards of IoT have been used by existing robotic systems and services?
 \item Which sensors and actuators have been incorporated in IoT-based robots?
 \item In which application areas has IoRT been applied?
  \item Which technologies of IoT have been used in robotics till today?
 \item Has WoT been used in robotics? If yes, by means of which technologies?
 \item Which is the overall potential of WoT in combination with robotics, towards a WoRT?
\end{enumerate}

As the IoT has been defined in different ways in literature, it must be clarified that IoT -in the context of this paper-
is perceived as a \textit{system that involves real-world things, which can communicate/interact over the Internet and they can be remotely monitored and controlled via Internet protocols}. These devices involve robots and robotic systems in the context of this work. 

Search for related work was performed through the Web scientific indexing services \textit{Web of Science} and \textit{Google Scholar}. The following query was used:\\
\textit{Robotics AND ["Internet of Things" OR "Web of Things"]}\\
Thirty seven (37) papers were found via this approach. To increase the range of our bibliography,
a search of the related work as appeared in these 37 papers was also performed. This effort allowed to increase
the number of papers discovered to 61. From these 61 papers, 12 papers (20\%) were discarded, as they did not involve a real-world implementation of a robotic system or because they have  not used  IoT or WoT somehow in their implementations. Forty nine (49) papers were finally selected in order to be analyzed in more detail. Each of these papers was studied in detail, aiming to address the aforementioned six research questions. The results of our research are presented in the next sections. 

\begin{figure}[h]
   \centering
   \vspace{-0.2cm}
   \includegraphics[width=0.60\columnwidth]{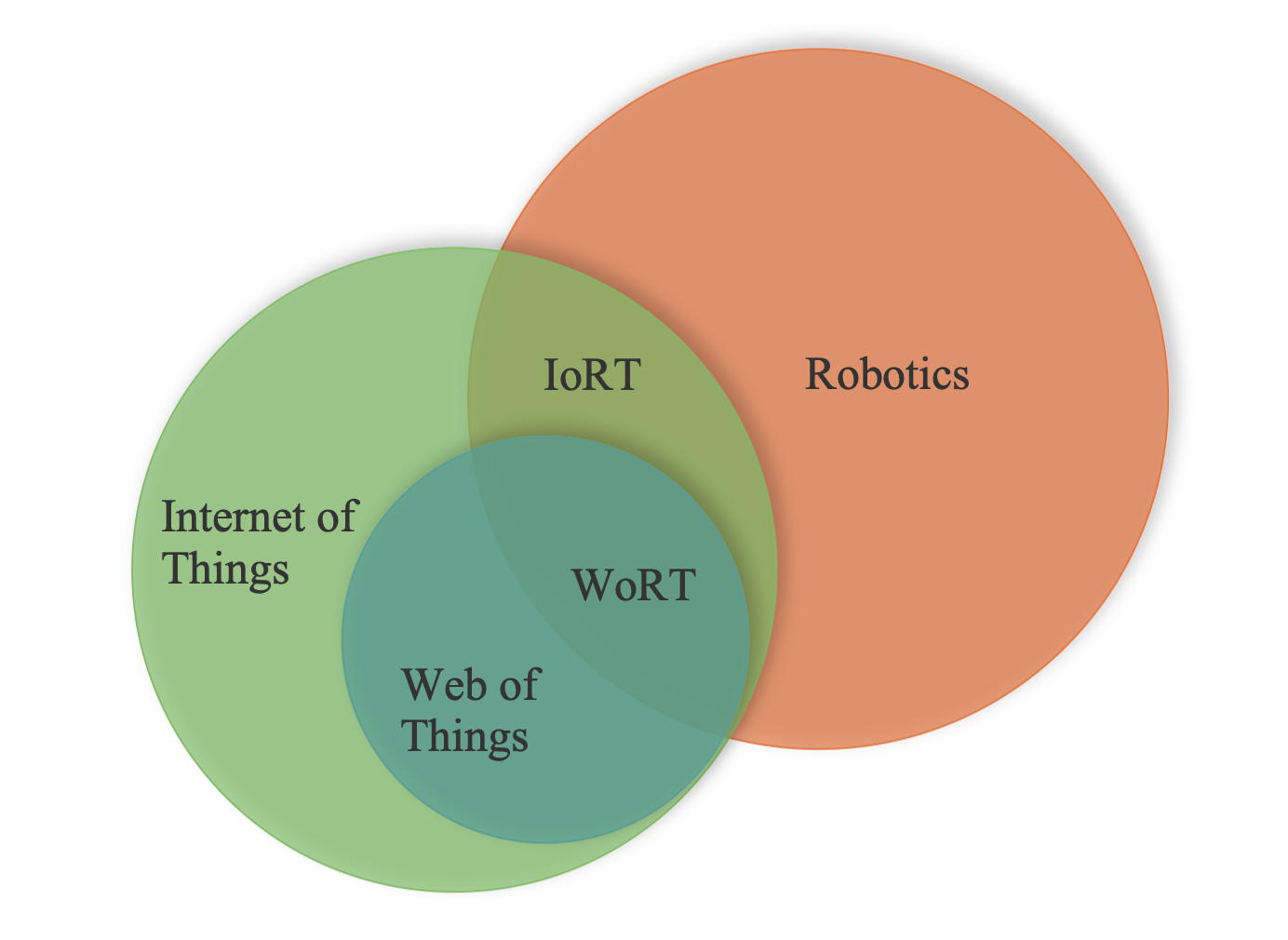}
   \caption{A Venn diagram showing research areas involved.}
   \label{fig:VennDiagram}
   \vspace{-0.3cm}
\end{figure}

We note that this paper focuses on the connection between robotics and the concepts/principles of IoT and WoT, aiming to close the existing gap in literature, as described in the introductory section. This paper does not intend to compare specific hardware and software platforms used for sensors/robots development and/or development of IoT-based systems. For such comparisons, the reader should consider relevant studies \cite{vermesan2017internet, mineraud2016gap, kamilaris2016web, chien2011comparative, gaur2015operating, rodrigues2010survey, pwcReport}, which cover quite well the spectrum of embedded hardware/software development, plus low-power communications. The research areas targeted by this paper are displayed as a Venn diagram in Figure \ref{fig:VennDiagram}. WoT is considered a subset of IoT and they both interact with robotics via the emerging research areas of IoRT and WoRT. Additional circles in the figure could be sensor technologies and communication protocols, as they are extensively used both by IoT and robotics. 

We note that IoRT is mostly about technical aspects of robotic systems, as well as technologies for communication and message exchange or services and data understanding. It is not about artificial intelligence, robot perception and empathic behaviour, where robotics touches upon other research disciplines.

\section{Robotics}
\label{robots}
Robots are becoming a fundamental part of our society and they will become even more important in future. The past decades were characterized by the massive automation in the industry, as for example in the case of automatic machines and industrial manipulators. In this context, the robots work in a perfectly known and modelled environment and safety layers are built around them to prevent harming people and other machinery. Furthermore, the actions of these robots are completely programmed in order to avoid any unpredictable behaviours. The robots work fast and accurately in order to improve the efficiency of industrial processes.

However, to further progress in the integration with everyday life, robots, usually referred as \textit{service robots} in this context, need to be able to perceive and understand unknown and complex environments and to be capable of planning and acting in unforeseen situations. In robotics, these concepts are usually gathered together under the word \textit{cognition}. Cognition is the key aspect in order to deal with the variety of environmental aspect, parameters and tasks in which the service robots operate, i.e. houses, warehouses and offices. The second major challenge is the manipulation of a high variety of objects, deformable or not, in the operational environment. The final hurdle for robots is the interaction. Robots and humans are expected to end up in the same environments, working in parallel and collaborating together, without any safety layer to keep them apart. Thus, the robots' behaviour must be predictable and safe for humans, for the surroundings and eventually for other co-operating robots. The concept of interaction, however, cannot be limited only to physical and safe interaction between humans and robots, but it must be extended to more abstract ways of interacting. Interpreting verbal and non-verbal communication such as as facial expression, body movements and gestures, as well as understanding of social interactions are clearly necessary steps to be taken in the near future. 

\subsection{Internet of Things and Robotics}
\label{iotrobbenefits}
\begin{figure*}[h]
   \centering
   \includegraphics[width=1.00\linewidth]{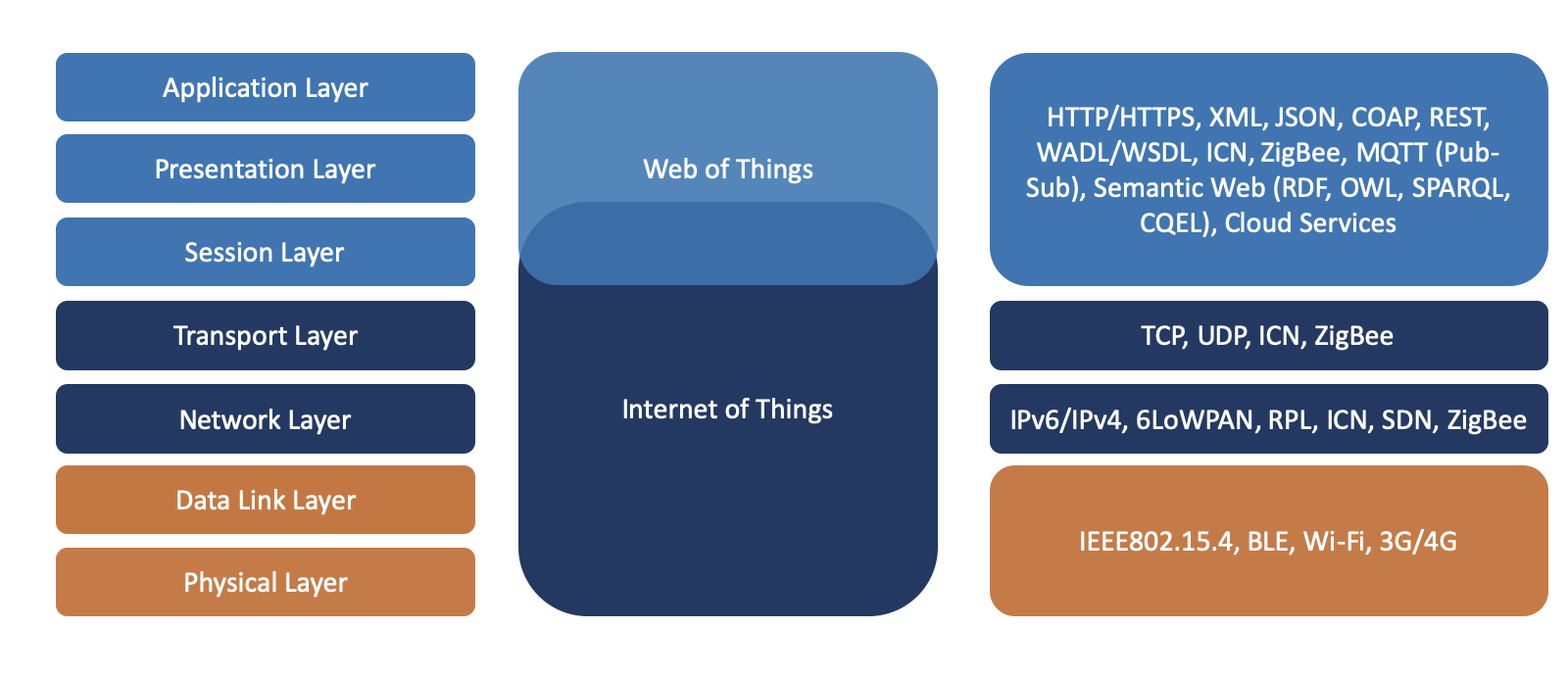}
   \caption{Layers of the ISO stack where IoT and WoT are located.}
   \label{fig:IoTdiffWoT}
\end{figure*}

Since robots are entering the everyday life of humans, supporting their tasks and interacting with them, the sensory equipment, hardware platforms, software and communication patterns of the robot machines become more complicated and demanding. As mentioned before, robots need to interact with their environment, i.e. other robots, devices and machines (R2R) and also with humans (R2H). To achieve this in a highly interoperable way towards true machine-to-machine (M2M) interaction involving context understanding, considering the wide variety of protocols and hardware/software/communication architectures and solutions available, the IoT and WoT come into play. The particular benefits from a WoT integration are listed in Section \ref{woTRbenefits} below. The main difference between IoT and WoT is that IoT operates in the lower layers of the ISO stack, while WoT mainly at the application layer. Some overlap between IoT and WoT may exist at the presentation and session layers of the ISO stack. This difference is illustrated in Figure \ref{fig:IoTdiffWoT}. The right part of the figure lists many of the technologies and acronyms mentioned through this survey, under the relevant part of the ISO stack where they belong, as well as whether they constitute mainly IoT or WoT technologies.

IoT could improve robotics with higher productivity (i.e. by re-using well-accepted and understood software and protocols), lower costs, better customer experiences due to the easier integration with existing components of the nearby environment and with cloud computing, high-quality data in terms of semantics awareness, context understanding and many other possibilities listed in related work \cite{weber2010internet, wilde2007putting, miorandi2012internet}. 

Moreover, robotics can benefit from the plethora of research and development in IoT, in terms of resource-constraint hardware and software, low-power communication algorithms and protocols, as well as optimal solutions for wireless sensor networks (WSN), such as networking, mobility, data propagation, topology building and maintenance etc. \cite{razafimandimby2016neural}. Finally, the unique naming and addressing capabilities provided by the IPv6 protocol, allow robots and robotic system to become uniquely addressable citizens of the Internet, exploiting the TCP/IP protocols for device discovery, message exchange, security etc.

\subsection{Web of Things and Robotics}
\label{woTRbenefits}
One of the first robots having a basic presence on the Web was MERCURY \cite{goldberg1995desktop}, which started its operation in 1994. It was one of the first tele-operated manipulators on the Web. It enabled Web users to excavate artefacts buried in a sand-filled terrarium. The TELE-GARDEN robot in 1995 \cite{goldberg1995telegarden}, successor of MERCURY, allowed people to control the planting of flowers via a Web interface, being able to coordinate requests by multiple users. XAVIER followed in 2002 \cite{simmons2002xavier}. Although it had only a basic Web interface allowing only basic interaction possibilities, it became quite successful due to the possibilities of remote control of a robot via any Web browser around the world.

As mentioned in the introduction, WoT could offer more benefits to robotic systems than IoT alone \cite{wilde2007putting, guinard2011internet}, especially in terms of interoperability at the application layer among robotic components and services, but also between robots (R2R) and humans (R2H), as well as among other machines (M2M). Some of these benefits are listed below, in more detail, in relation to robotic systems:
\begin{itemize}
 \item Data from the sensors of the robots may be easily exported into Web applications in popular, well-understood standard formats, for easier reuse. Representation formats may be negotiated in real-time, depending on the formats supported by the machines involved.
 \item Exposing the services provided by the robots (and their individual sensing and actuating components) as interoperable application programming interfaces (APIs), would provide the primitives to users with little programming experience to perform advanced tasks. Users may select any programming language that supports the HTTP protocol, such as Python, Java, Ruby, C, PHP, JavaScript etc.
 \item Exposing robotic services as APIs would facilitate application-layer interoperability between robots (R2R) and humans (R2H). When these APIs become standardized, enhanced with semantic technologies, interaction between robots and humans can become automatic and in real-time, using the Web as the common platform and language for communication.
 \item Combining robotic services with Web services and resources would allow the creation of Web-based robotic mashups, where the robots exploit seamlessly knowledge, information and context already available at the Web, towards more informed choices and aware behaviour.
 \item Uniform access to heterogeneous embedded devices installed on the robot, where the robot itself becomes a homogeneous environment (i.e. at the application layer), where any sensor/actuation can be individually accessed in a standardised way, facilitating coordination, action-taking and control. This homogeneous access would allow easier connection between robotic systems and cloud computing, for more advanced real-time computing processing and for permanent storage of information such as sensory measurements.
 \item Harnessing well-defined protocols used for years on the Web for device and service discovery, service and data description, semantics understanding, security and privacy, orchestration and routing.
 \item Particularly related to semantics of services and information, the Semantic Web (see Section \ref{webSemant} below) involves numerous technologies and implementations for uniformly describing devices, services and data, allowing common understanding and advanced reasoning between different entities.
\end{itemize}

We note that APIs in the WoT research area are expected to be resource-oriented and to follow the 
REpresentational State Transfer (REST) \cite{feng2009rest}. REST is an architectural style for developing Web Services and it has been adopted by WoT due to its simplicity and the fact that it builds upon existing systems and features of the HTTP protocol. A RESTful API follows the principles of REST for providing Web Services to the users, based on the HTTP request/response protocol. Thus, basic HTTP-based interfaces that do not follow REST can not be considered part of the WoT. However, they can still be considered part of the IoT, as long as they employ TCP/IP. 

\section{Analysis of the Application of Internet of Things in Robotics}
\label{IoT}
This section addresses the first three research questions as defined in Section \ref{meth}.
First, Section \ref{IoTapps} lists the existing applications of IoT in robotics,
presenting how sensors and robot actions have been used in different application domains.
Then, Section \ref{IoTplatforms} shows some of the hardware elements and sensory equipment used in the work under study. Afterwards, Section \ref{IoTsoftCom} presents the software platforms and communication protocols used during programming and control of IoT-based robotic systems. Finally, Section \ref{IoTrelRobots} summarizes in which ways the IoT-enabled robots of related work have made actual use of Internet technologies.

\subsection{Applications}
\label{IoTapps}
Table \ref{tab1} lists the different application domains and their specific application areas, where IoT-based robots and robotic services have been used. Most popular categories are entertainment (8 papers), health (7 papers), education (6 papers), surveillance (6 papers) and culture (5 papers). Some categories are overlapping, such as health with domestic support, surveillance with military, as well as emergency/disaster response with rescue operations. Autonomous cars are used as moving robots in the area of transportation, while a warfare robot car has been developed for military purposes in \cite{imran2018design}.
Unmanned aerial vehicles (UAV) (i.e. drones) are considered as flying robots in surveillance, emergency/disaster response and rescue operations.

\begin{figure*}
\begin{subfigure}{.31\textwidth}
  \centering
  \includegraphics[width=0.52\linewidth]{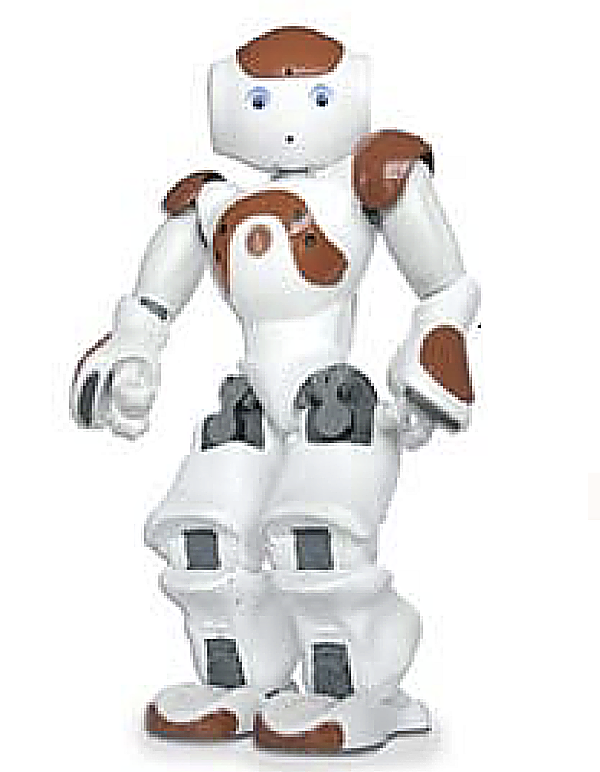}
  \caption{Robot used for diabetes management \\for children (Source: \cite{al2017robot}).}
\end{subfigure}%
\begin{subfigure}{.31\textwidth}
  \centering
  \includegraphics[width=1.01\linewidth]{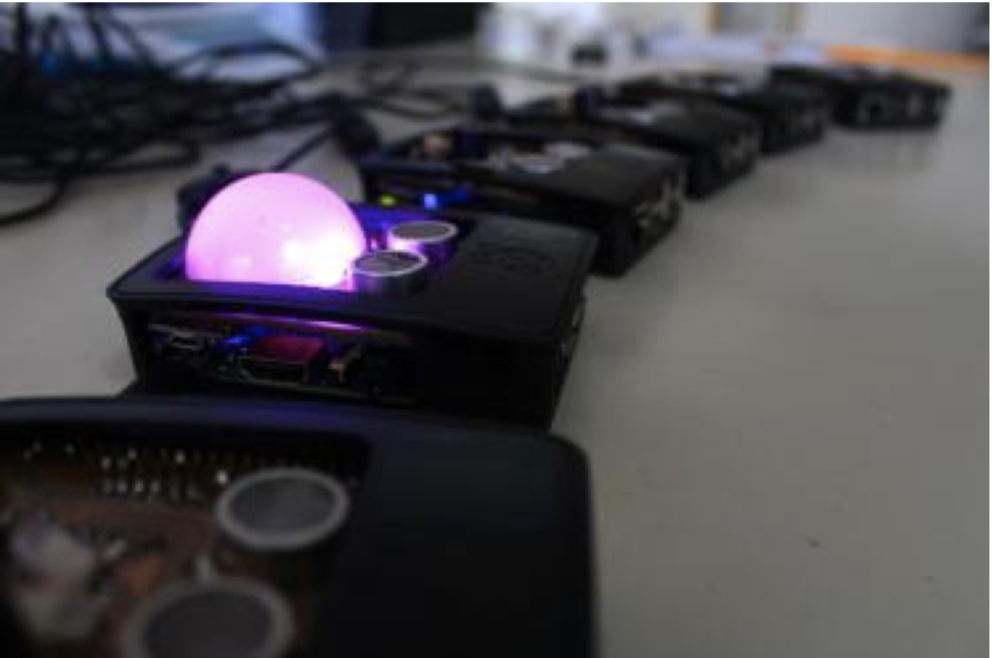}
  \caption{Robot for measuring people's reflexes (Source: \cite{doxopoulos2018creating}).}
  \end{subfigure}
  \begin{subfigure}{.31\textwidth}
  \centering
  \includegraphics[width=0.95\linewidth]{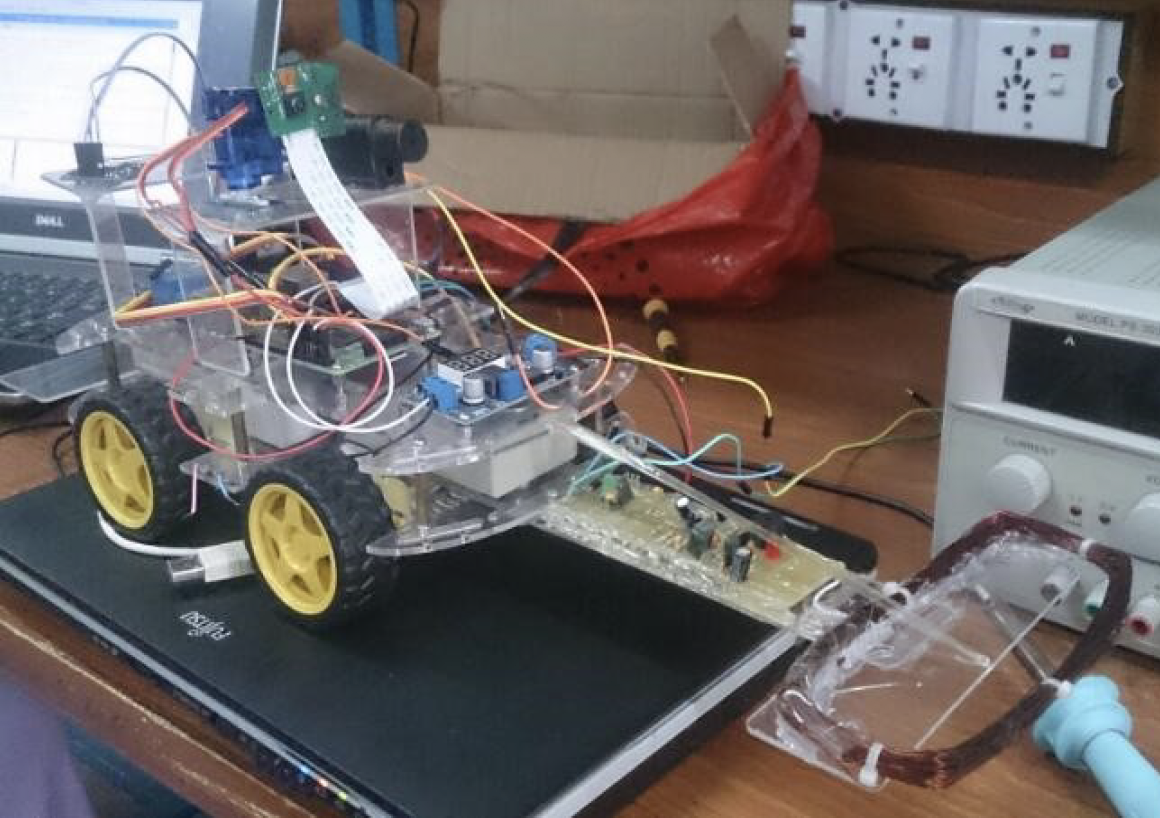}
  \caption{Robot car model used for warfare (Source: \cite{imran2018design}).}
\end{subfigure}
\newline
\vspace{0.4cm}
\newline
\begin{subfigure}{.31\textwidth}
  \centering
  \includegraphics[width=0.70\linewidth]{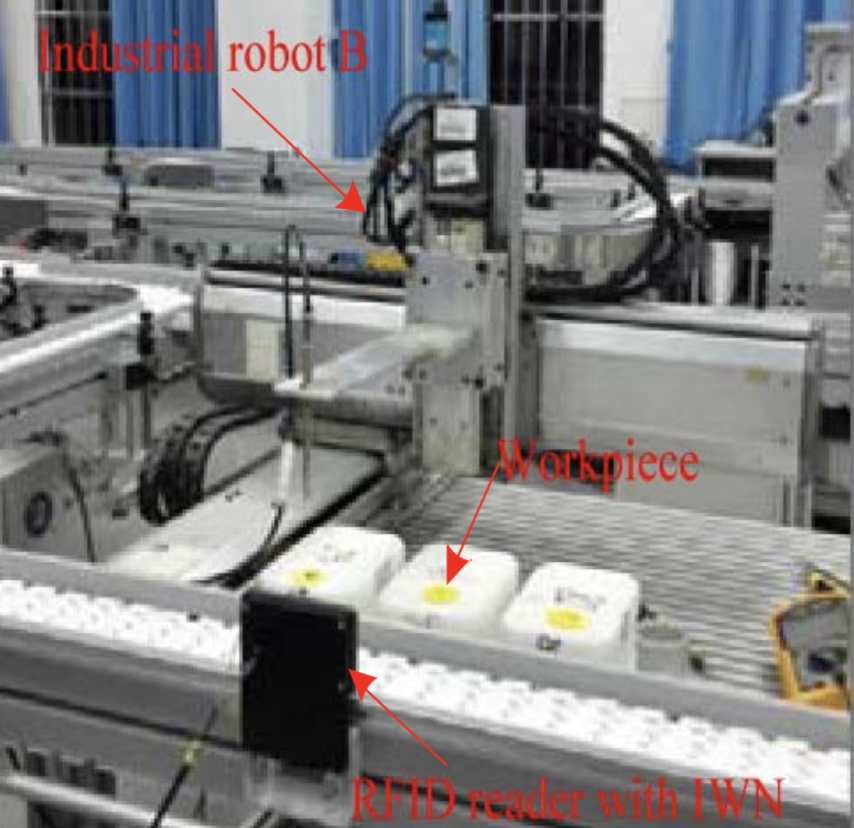}
  \caption{Robot used for material handling at \\the industry (Source: \cite{wan2016software}).}
\end{subfigure}%
\begin{subfigure}{.31\textwidth}
  \centering
  \includegraphics[width=0.90\linewidth]{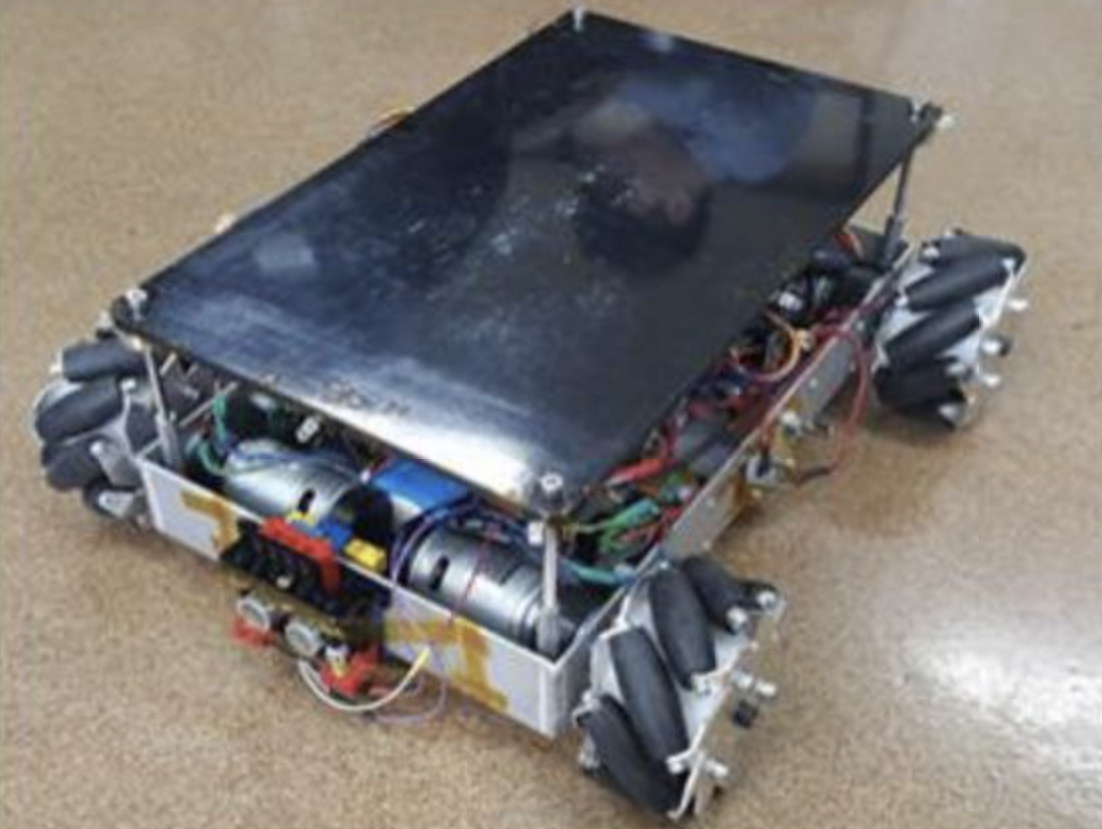}
  \caption{Robot presenting pet behavior \\(Source: \cite{kim2016self}).}
  \end{subfigure}
  \begin{subfigure}{.31\textwidth}
  \centering
  \includegraphics[width=0.87\linewidth]{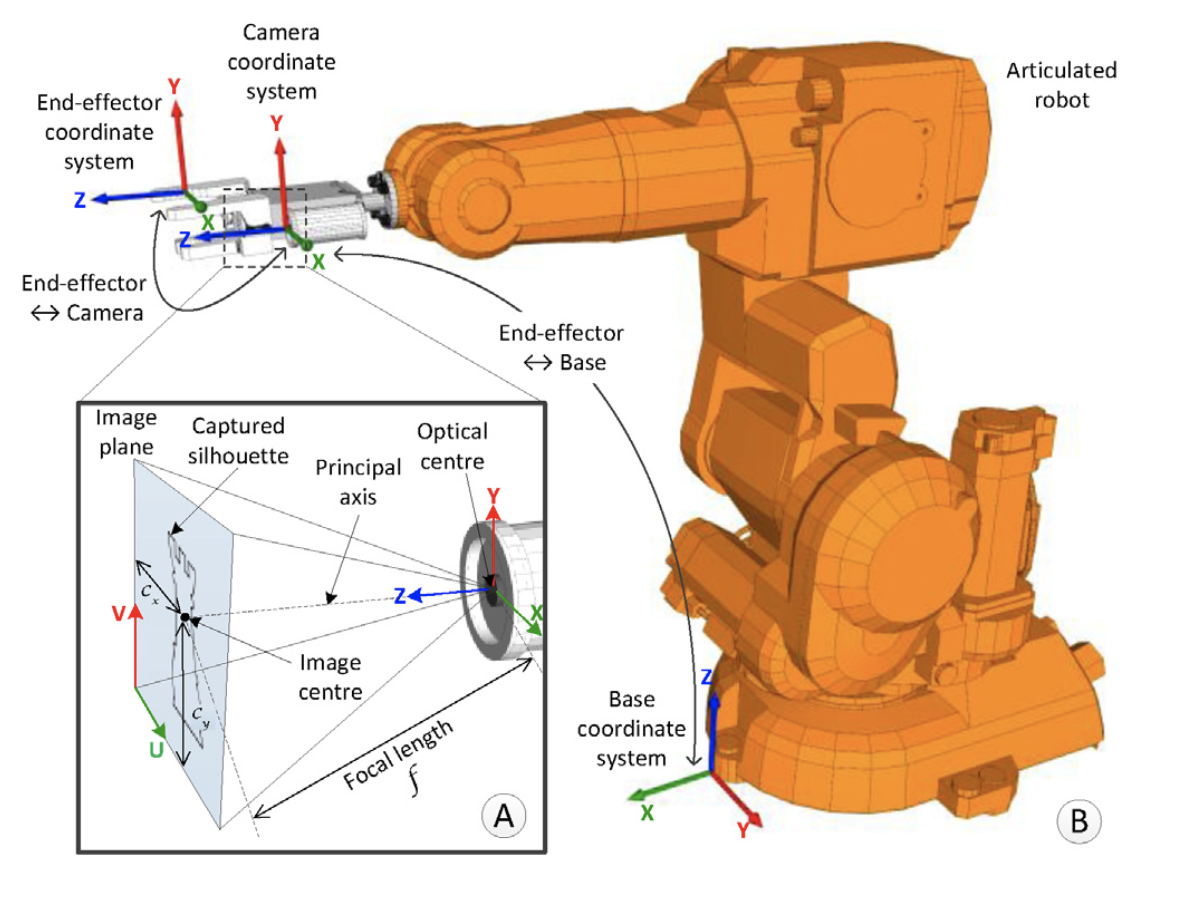}
  \caption{Robot used in agriculture for crops' monitoring(Source: \cite{wang2014remote}).}
\end{subfigure}
\caption{Example robots as presented in related work under study. }
\label{fig:examplerobots}
\end{figure*}

Example robots as they appear in different application areas of the work under study are shown in Figure \ref{fig:examplerobots}.

\begin{table*}[htbp]
\caption{Applications of IoT in robotics.}
\begin{center}
\begin{tabular}{|p{5.1cm} | p{11.9cm}|}
\hline
\textbf{Application Area} &  \textbf{Application}\\
\hline
Industry  & Manufacturing \cite{brizzi2013bringing, wan2016software}, material handling \cite{wan2018context}, 3D assembly operations \cite{wang2014remote} \\
\hline
Customer support  & Office operations' support \cite{mizoguchi1999human, simmons2002xavier} \\
\hline
Transportation  & Autonomous cars \cite{gerla2014internet}, car parking system \cite{ji2014cloud} \\
\hline
Environment  & Water quality monitoring \cite{wang2015new}, smoke detection \cite{Bhutada2018}, air quality \cite{Valluri2017}, space exploration \cite{backes2000internet} \\
\hline
Health  & Diabetes management \cite{al2017robot, al2015mobile, al2013web}, measuring reflexes \cite{doxopoulos2018creating}, tele-surgery \cite{cenk2003robotics}, tele-echography \cite{masuda2001three}, remote treatment \cite{takanobu2000remote} \\
\hline
Education  & Teaching computing, programming and robotics in schools and universities \cite{callaghan2012buzz, hamblen2013embedded, turcu2012integrating}, collaborative learning \cite{plauska2014educational}, educate people in public places \cite{burgard1999experiences} \\
\hline
Domestic support  & Medical and health care \cite{hu2013application}, support of people with dementia \cite{simoens2016internet}, support of elderly and disabled residents \cite{jia2002internet}, independent living \cite{coradeschi2013giraffplus} \\
\hline
Entertainment  & Pet robot \cite{kim2016self, luo2001multisensor}, remote painting \cite{stein1998painting}, robot that sings and dances \cite{lu2013robot}, entertain people in public places \cite{burgard1999experiences}, allow low-cost public access to a tele-operated robot \cite{goldberg1995desktop}, interact with a remote garden filled with living plants \cite{goldberg1995telegarden}, moving in a wooden labyrinth trying to get out of the maze \cite{saucy2000khepontheweb}  \\
\hline
Sports  & Ball detection and catching \cite{tanaka1999internet}, measuring reflexes \cite{doxopoulos2018creating}, control of a football robot team \cite{bi2017iot} \\
\hline
Culture  & Remote tour guiding at a museum \cite{maeyama2001remote, paulos1996delivering, burgard1999experiences}, interactive tour guide in museums \cite{thrun1999minerva, burgard1999experiences} \\
\hline
Surveillance  & UAV \cite{mozaffari2016mobile, mozaffari2017mobile, scilimati2017industrial}, monitoring activities in factories, offices and industrial sites, remote control \cite{shin2016design, scilimati2017industrial}, detect internal condition of working areas \cite{Valluri2017}, drones as a service \cite{loke2015internet} \\
\hline
Military  & Warfare robot car \cite{imran2018design}, land mining and field surveillance \cite{Ashokkumar2018} \\
\hline
Emergency/Disaster response  & Emergency response system \cite{zander2015cyber, harbertinternet}, drones for emergency use \cite{loke2015internet}, crime situations \cite{ermacora2013cloud} \\
\hline
Rescue operations & UAV for object tracking \cite{koubaa2018dronetrack}, UAV for disaster rescue operations \cite{ahn2018reliable} \\
\hline
Agriculture  & Soil moisture sensing and remote crop monitoring \cite{wang2015new}, live streaming of crops, seed sowing, pesticide sprinkling and automatic irrigation \cite{Hemalatha2018} \\
\hline
\end{tabular}
\label{tab1}
\end{center}
\end{table*}

\begin{figure*}[htbp]
\label{robotSensorsApps}
\centerline{\includegraphics[width=1.0\linewidth]{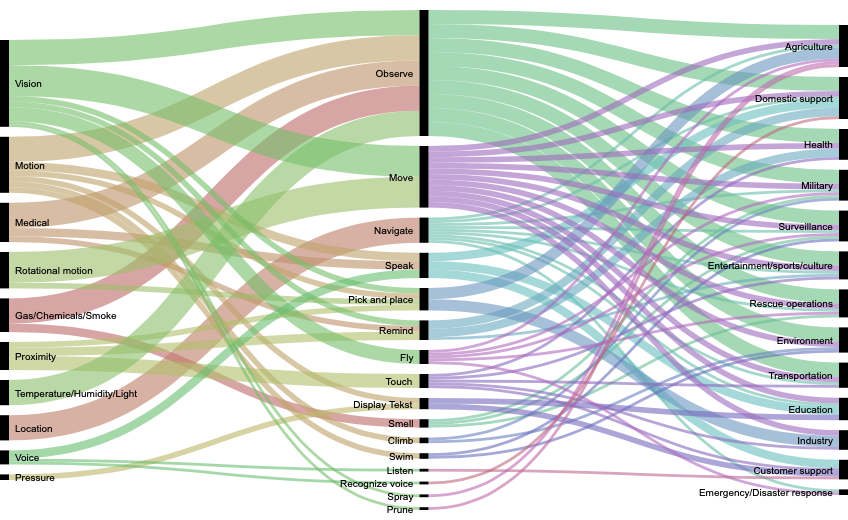}}
\caption{The connection between IoT robot sensors, actions and IoT application areas, as they appear in related work under study.}
\label{fig1}
\vspace{-0.5cm}
\end{figure*}

Figure \ref{fig1} maps together the sensing/actuation capabilities of the IoT robots, together with their actions, in relationship to the different application areas \textit{as encountered in this study by analyzing the related bibliography}. The most popular sensors, actions and application areas are highlighted in blue color. As the figure shows, the most commonly used sensors are the ones that measure proximity (for the actions of touching, picking and placing, reminding), vision (for the actions of picking and placing, moving, observing, flying, spraying and pruning, reminding), voice (for the actions of speaking, listening and voice recognition) and motion (for the actions of picking and placing, speaking, observing, climbing, swimming and displaying text). 

The most popular actions involve moving, observing, flying and navigating. Some actions are appropriate only for specific application areas, such as spray and prune for agriculture, reminders for health and domestic support, pick-and-place for customer support and industrial applications etc.

In relation to the possible actions performed by the robot, based on its sensory equipment, the application areas where most actions have been recorded are agriculture (7 actions), domestic support (6 actions), surveillance (6 actions), military (6 actions), rescue operations (5 actions) and entertainment (5 actions).

\subsection{Hardware and Sensors}
\label{IoTplatforms}

\subsubsection{Hardware}
Besides the mechanical parts which vary significantly per related work under study,
there is some common hardware used among the surveyed papers, such as Raspberry Pi \cite{wan2016software, Bhutada2018, Valluri2017, doxopoulos2018creating, callaghan2012buzz, shin2016design, imran2018design, koubaa2018dronetrack, Hemalatha2018}
and Arduino \cite{wang2015new, shin2016design, Ashokkumar2018}. 
Both Raspberry Pi and Arduino constitute open-source hardware and electronic prototyping platforms, enabling users to create interactive electronic objects. In the context of this survey, researchers in related work have used them as mini computers, connecting external sensors, actuators and mechanical parts, in order to give intelligence to their robots.
TelosB was the sensor platform selected in \cite{scilimati2017industrial}.

Some efforts tried to develop humanoid robots \cite{doxopoulos2018creating, simoens2016internet, burgard1999experiences, thrun1999minerva}, while others employed autonomous vehicles \cite{wan2016software, gerla2014internet, wang2015new}, a space exploration rocket \cite{backes2000internet}, a warfare car robot \cite{imran2018design} and UAV/drones \cite{mozaffari2016mobile, mozaffari2017mobile, loke2015internet, wang2015new, zander2015cyber, koubaa2018dronetrack, ahn2018reliable, ermacora2013cloud, scilimati2017industrial}. An UAV enhanced with IoT sensors has been introduced in \cite{scilimati2017industrial}, where the UAV interacts better with its environment towards more effective surveillance, gathering data coming from temperature, humidity and light sensors.

The remaining papers used robots with application-specific mechanics and characteristics. The interesting concept of \textit{biobots} was introduced in \cite{harbertinternet}, which was about real animals (in this case, dogs) equipped with sensors such as cameras or gas detectors. These sensors allowed the animals to sense additional environmental aspects (i.e. search rubble for casualties or detect dangers such as a gas leak). In this case, the mechanical parts of a robot are replaced by the physical capabilities of the animals involved.

\subsubsection{Sensors}
Robots were equipped with a wide variety of different sensors, such as RFID \cite{brizzi2013bringing, wan2016software, wan2018context, wang2015new, hu2013application}, video cameras \cite{mizoguchi1999human, saucy2000khepontheweb, wang2014remote, gerla2014internet, backes2000internet, plauska2014educational, jia2002internet, tanaka1999internet, goldberg1995desktop}, infrared and light sensors \cite{mizoguchi1999human}, smoke sensors \cite{Bhutada2018}, temperature and gas \cite{Valluri2017, Ashokkumar2018}, temperature, humidity and light sensors \cite{scilimati2017industrial}, medical sensors \cite{al2015mobile, al2017robot, hu2013application}, accelerometers and gyros \cite{hamblen2013embedded, kim2016self}, occupancy \cite{coradeschi2013giraffplus} and infrared sensors \cite{Hemalatha2018, mizoguchi1999human, saucy2000khepontheweb}.
GPS receivers were also installed in many robots for localization and navigation \cite{gerla2014internet, wang2015new, hamblen2013embedded, loke2015internet, imran2018design, koubaa2018dronetrack, ermacora2013cloud}.

The relationship between the aforementioned sensors and the robots' sensing/actuation capabilities, as they appear in related work, is presented in Table \ref{SensorsActions}. The relationship between the intended use of these sensors, the desired actions and targeted application areas is illustrated in Figure \ref{fig1}.

\begin{table}[htbp]
\caption{Sensors and intended use.}
\begin{center}
\begin{tabular}{|p{3.5cm} | p{4.5cm}|}
\hline
\textbf{Sensor} &  \textbf{Intended Use} \\
\hline
Microphone & Voice recognition \\
\hline
Gyrometer, accelerometer & Rotational motion \\
\hline
RFID, Infrared sensor & Sense proximity \\
\hline
Occupancy, infrared sensor & Sense motion, perform some mechanical action \\
\hline
Video camera & Computer vision, human remote vision \\
\hline
Pressure sensor & Sense pressure on something \\
\hline
Medical sensors & Measure health indications of humans \\
\hline
GPS receiver & Location \\
\hline
Gas/chemical sensor & Sense hazardous materials \\
\hline
Smoke sensor & Sense fire \\
\hline
Temperature/humidity sensor & Sense weather conditions, surveillance \\
\hline
Light sensor & Measure illumination, surveillance \\
\hline
\end{tabular}
\label{SensorsActions}
\end{center}
\end{table}

\subsection{Software and Communications}
\label{IoTsoftCom}

\subsubsection{Software}
\label{software}
In regards to software, some papers \cite{koubaa2018dronetrack, ahn2018reliable, ermacora2013cloud, scilimati2017industrial} employed the popular Linux-based Robot Operating System (ROS) \cite{quigley2009ros}, which provides the communications infrastructure to program, operate, debug and control the robot as a system of systems.

ROS is an open-source framework for writing software for robotic systems. It consists of a number of libraries, tools and sets of conventions to simplify the task of writing software for complex mechatronic systems \cite{quigley2009ros}. ROS includes a large variety of algorithms and functions for creating new software components and drivers. ROS supports multiple sensor technologies as well as programming languages, of which C++ and Python are the most important. 
Nowadays, ROS is one of the most exploited tools for developing algorithms in the context of robotics due to its flexibility and simplicity. However, it is worth mentioning that ROS has no real-time capabilities. This means that ROS does not provide guarantees about the timing of operations, hence it is not intended for operations that have strict timing requirements.

In the case of safety critical systems in which hard real-time constraints exist, real-time operating systems must be adopted.
The real-time operating system (RTOS) is one such system, used in  \cite{hamblen2013embedded}, which handles the execution of tasks in order to meet their time deadline, but also facilitates memory management and accessing resources.  The two main design philosophies are: event-driven and time sharing. An event-driven scheduler switches between tasks when an event of higher priority requires to be accommodated. On the other hand, a time sharing scheduler switches among tasks based on a periodic clock signal.

Other papers used GOLEX \cite{burgard1999experiences, thrun1999minerva}, Embedded C \cite{Ashokkumar2018}, the Multi-target Robot Language (MRL) \cite{mizoguchi1999human}, Node.js \cite{shin2016design} as well as OpenWSN \cite{scilimati2017industrial}. OpenWSN  
is an open-source implementation of the IEEE/IETF 6TiSCH protocol stack \cite{watteyne2012openwsn}. 6TiSCH is a promising Working Group that aims to achieve industrial-grade performance in terms of jitter, latency, scalability, reliability and low-power operation for IPv6 over IEEE802.15.4e TSCH.

Besides the aforementioned software systems and platforms, Ray \cite{ray2016internet} has listed numerous emerging cloud-based robotics platforms that could be used together with IoRT architecture.
As an example, Simoens et al. \cite{simoens2016internet} employed the DYAMAND middleware \cite{nelis2012dyamand} for abstracting the different protocols and interfaces of the installed sensors of the robot.

\subsubsection{Communication}
In terms of communication protocols, the majority of related work used Wi-Fi (i.e. 16 papers), while IEEE802.15.4 and ZigBee \cite{imran2018design, wan2016software, wan2018context, hu2013application, scilimati2017industrial}, 3G/4G \cite{al2015mobile, al2017robot, koubaa2018dronetrack} and Bluetooth \cite{al2015mobile, al2017robot, hu2013application, simoens2016internet, kim2016self, imran2018design, koubaa2018dronetrack} were also used. Regarding Bluetooth in particular, the Bluetooth Low Energy (BLE) technology was used for communication among robot components.

Wi-Fi has been used where wide coverage was required (i.e. up to 100 meters) and/or it was not possible to propagate messages via intermediate nodes \cite{al2017internet}. Since Wi-Fi consumes much energy, it has been used where autonomy had not been an important issue (e.g. education, health, tour guiding in museums). 3G/4G has also been used in scenarios where connectivity was difficult (e.g. rescue operations) but it consumes more energy in comparison to Wi-Fi. On the other hand, ZigBee and BLE have been used for short-range and low-energy communication scenarios, where there was a specific (indoor) topology/infrastructure and need for autonomy (e.g. industry, domestic support, entertainment).

An aspect not really addressed in the surveyed papers is the security of software, communication messages and the actual data involved. Two papers mentioned the use of the HTTPS protocol for secure communication \cite{al2015mobile, al2017robot} while NASA employed the WITS Encrypted Data Delivery System (WEDDS) and its public key infrastructure during the Mars polar lander mission \cite{backes2000internet}.

\subsection{Summary}
\label{IoTrelRobots}
It is worth investigating in which ways the IoT-enabled robots of the related work under study make actual use of Internet technologies, according to the definition of IoRT \cite{ray2016internet, vermesan2017internet, simoens2018internet}, as mentioned in the introduction. Table \ref{tab2} presents the classification of the surveyed papers in different classes, based on how they use Internet technologies.

\begin{table*}[htbp]
\caption{Use of Internet technologies in IoRT.}
\begin{center}
\begin{tabular}{|p{6.3cm} | p{7.7cm}|}
\hline
\textbf{Type of use} &  \textbf{Related Work}\\
\hline
Not any details  & \cite{kim2016self, mozaffari2016mobile, mozaffari2017mobile, ahn2018reliable} \\
\hline
IPv4 communication & \cite{wang2015new, saucy2000khepontheweb, Bhutada2018, Valluri2017, backes2000internet, al2015mobile, al2013web, doxopoulos2018creating, takanobu2000remote, callaghan2012buzz, hamblen2013embedded, plauska2014educational, hu2013application, simoens2016internet, coradeschi2013giraffplus, luo2001multisensor, stein1998painting, lu2013robot, tanaka1999internet, bi2017iot, burgard1999experiences, maeyama2001remote, thrun1999minerva, shin2016design, imran2018design, paulos1996delivering, simmons2002xavier, goldberg1995desktop, goldberg1995telegarden} \\
\hline
Java Object Request Broker (ORB) architecture & CORBA \cite{jia2002internet}, HORB \cite{masuda2001three} \\
\hline
IPv6 architectures & 6LoWPAN \cite{brizzi2013bringing}, 6TiSCH technology (COAP, RPL, 6LoWPAN) \cite{scilimati2017industrial} \\
\hline
Cloud robotics & \cite{ermacora2013cloud, loke2015internet, wan2016software, wan2018context, gerla2014internet, Ashokkumar2018, zander2015cyber, koubaa2018dronetrack, Hemalatha2018, ermacora2013cloud} \\
\hline
\end{tabular}
\label{tab2}
\end{center}
\end{table*}

The majority of the surveyed papers use TCP/IP communications for interaction between the robot and the outside world \cite{rodrigues2010survey}. Some of these papers incorporate principles of the WoT and they would be analyzed in the Section \ref{WoT}. Java ORB is used in \cite{masuda2001three, jia2002internet} for managing distributed program objects, while Brizzi et al. \cite{brizzi2013bringing} employ an IPv6-based 6LoWPAN architecture for communication between robots and wireless sensor networks. 6LoWPAN is a working group and standard for the application of IPv6 over low-power sensors and wireless sensor networks \cite{mulligan20076lowpan}. An implementation of 6LoWPAN via the 6TiSCH technology  \cite{watteyne2012openwsn} was performed in \cite{scilimati2017industrial}, combining 6LoWPAN with relevant software protocols and implementations such as the Constrained Application Protocol (COAP) \cite{shelby2014constrained} and the RPL IPv6 routing protocol \cite{winter2012rpl}.

Finally, the last category of Table \ref{tab2} is about papers using cloud services for storage, processing, updates and management/control. These papers touch upon the research area of \textit{cloud robotics} \cite{hu2012cloud}, which deals with infrastructures and protocols for machine-to-cloud (M2C) communications. Some interesting relevant concepts mentioned in related work \cite{gerla2014internet, wan2016software} are Information Centric Networking (ICN) \cite{ahlgren2012survey} and Software-Defined Networking (SDN) \cite{mckeown2009software}. ICN is an approach to evolve the Internet infrastructure away from a host-centric paradigm based on the end-to-end principle, to a network architecture in which information is the focal point. SDN is an architecture that aims to make networks agile and flexible, improving network control. In the context of IoRT, ICN and SDN are approaches/architectures for facilitating robot control, as well as communication and networking between robots and the outside world through the Internet.

\section{Analysis of the Application of Web of Things in Robotics}
\label{WoT}
Table \ref{tab3} lists which Web technologies have been used in the surveyed papers. Papers that do not use any Web technologies have been omitted from the table. 
Most of the papers use a basic HTTP interface for interacting with the robot, while the underlying communication is realized using different communication protocols. In Section \ref{woTRbenefits}, it was mentioned that WoT-based developments should employ REST and not only basic HTTP interfaces. However, we still include in Table \ref{tab3} papers with only basic HTTP interfaces for interaction with robots, because they constitute the majority of related work.

Actual Web servers on the robot are installed in \cite{hamblen2013embedded} (i.e. commercial Keil Tools C/C++ cloud-based compiler) and \cite{shin2016design} (i.e. NodeJS server platform), while some papers move one step further, creating RESTful APIs for interacting with the robot's features and operations \cite{Valluri2017, doxopoulos2018creating, koubaa2018dronetrack}. 

To enable Web-based interaction, some papers used platforms such as the WebIOPi IoT framework \cite{Valluri2017}, the Web interface for telescience (WITS) \cite{backes2000internet}, the WAMP server \cite{doxopoulos2018creating}, the MASSIF platform \cite{bonte2017massif, simoens2016internet} and NodeJS \cite{shin2016design}. 
These platforms provide Web servers and support for exposing robot services as API calls, control and debugging of the software used for programming the robots (e.g. WebIOPi for Raspberry Pi), control dashboards and easy installation setups (e.g. WAMP), remote access to the robot (e.g. WITS for the planetary rover mission to Mars \cite{backes2000internet}), management of asynchronous messaging and handling of thousands of concurrent connections (e.g. NodeJS), semantic annotation, reasoning and integration of IoT data (e.g. MASSIF \cite{bonte2017massif}) etc.

Shin et al. \cite{shin2016design} used the Web Service Description Language (WSDL) to describe the services provided by the REST API of their UAV \cite{christensen2001web}. WSDL \cite{christensen2001web}, similar to Web Application Description Language (WADL) which is generally more suitable for Web-based applications \cite{hadley2006web}, is used to describe service and their semantics in order for humans and machines to be able automatically to use these services by creating the appropriate requests via TCP/IP calls.

Finally, Web-based message brokers, such as RabbitMQ and Crossbar.io, are used in \cite{simoens2016internet, doxopoulos2018creating}. These Web messaging brokers are suitable for high-performance, scalable distributed messaging where multiple publishers and subscribers of information are involved (e.g. health monitoring scenarios with care-providers involved \cite{simoens2016internet, doxopoulos2018creating}).

\begin{table}[htbp]
\caption{Use of Web technologies in WoRT.}
\begin{center}
\begin{tabular}{|p{3.8cm} | p{4.2cm}|}
\hline
\textbf{Type of use} &  \textbf{Related Work} \\
\hline
HTTP-based communication & \cite{ermacora2013cloud, scilimati2017industrial} \\
\hline
Basic HTTP interface for interacting with the robot  & \cite{goldberg1995desktop, goldberg1995telegarden, simmons2002xavier, mizoguchi1999human, saucy2000khepontheweb, wang2015new, Bhutada2018, backes2000internet, al2015mobile, al2013web, jia2002internet, stein1998painting, burgard1999experiences, maeyama2001remote, thrun1999minerva, Ashokkumar2018, Hemalatha2018, paulos1996delivering} \\
\hline
Web server on the robot & \cite{hamblen2013embedded, shin2016design} \\
\hline
REST API for robot control & \cite{Valluri2017, doxopoulos2018creating, koubaa2018dronetrack} \\
\hline
Semantic Web technologies &  \cite{simoens2016internet, brizzi2013bringing} \\
\hline
Publish/Subscribe architectures (Message brokers) & \cite{simoens2016internet, doxopoulos2018creating} \\
\hline
\end{tabular}
\label{tab3}
\end{center}
\end{table}

\subsection{Web Semantics}
\label{webSemant}
An \textit{advanced} aspect of WoT towards seamless M2M communication and understanding is the use of Semantic Web technologies
to describe services and data, towards a semantic WoT \cite{pfisterer2011spitfire}. The semantic WoT involves technologies for uniformly describing WoT data streams, devices and services, allowing easy and fast integration with other sources of information, platforms and applications towards advanced knowledge and reasoning \cite{Kamilaris2017a}.

Semantics are particularly important in robotics, for
the understanding of space, ambient environment and surroundings of the robot and for better reasoning. Semantics can be described and analyzed by means of ontologies and vocabularies (e.g. OWL, SSN \cite{compton2012ssn}, IoT \cite{seydoux2016iot}). Ontologies include concepts and categories in a subject area or domain (i.e. where the robot is operating) that show, describe and explain their properties and the relations between them. Ontologies are not enough though and need to be accompanied by description languages (e.g. RDF, OWL), as well as query languages (e.g. SPARQL, CQEL). Description languages facilitate consistent encoding, exchange and processing of semantically-annotated content.
The Resource Description Framework (RDF) is a general-purpose language for representing information on the Web, while the Web Ontology Language (OWL) is a formal language for representing ontologies in the Semantic Web. Semantic query languages are necessary for retrieving and manipulating data stored in description languages such as RDF/OWL, being able to answer complex queries and produce advanced knowledge by combining different sources of information together.

From the surveyed papers, Simoens et al. \cite{simoens2016internet} employed semantic technologies to describe the robots' produced data by means of the WSN and SSN ontologies while Brizzi et al. \cite{brizzi2013bringing} described robotic services by means of semantic web services. Semantic Web services \cite{mcilraith2001semantic} are similar to Web Services, but they additionally employ standards for the interchange of semantic data.

\section{Discussion}
\label{Disc}
This section discusses the general findings of the survey. Specifically, Section \ref{technicalaspects} refers to the technical aspects of the surveyed work and Section \ref{linkIoT} examines the actual relationship between related work under study and the general principles of IoT/WoT. Then, Section \ref{bigPicture} captures the big picture in relation to research and development in the area of robotic IoT, while Section \ref{future} provides future research opportunities in the IoRT domain. Finally, Section \ref{DemoPIRATE} demonstrates our own IoT-enabled robot, used for inspection of pipelines and Section \ref{takeAwayMessages} summarizes the take-home messages of this survey paper.

\subsection{Technical Aspects}
\label{technicalaspects}
A large percentage of the research work under study (36\%) employed open-source hardware and prototyping platforms, connecting them to a wide variety of sensors (i.e. 14 different sensor types). The mechanical parts of the robots involved 16 different actions, with movement, observation and navigation being the most popular ones. Sixteen application areas have been recorded, with health being the most popular application domain for IoT-based robotics. This makes sense, considering that health applications, especially at domestic level, require easy interconnection with other devices of a smart home or fast notification/alerting and communication with care providers. Thus, the IoT/WoT protocols are appropriate for this interconnection with low effort.

In regards to communications, Wi-Fi was the most popular technology (36\% of the papers), followed by Bluetooth (16\%) and ZigBee (10\%). The characteristics of these communication protocols are described in \cite{al2017internet} and researchers should consider the most appropriate technology for their implementations, taking into account application requirements such as range and coverage, energy consumption and autonomy of the robot, security, mobility aspects etc.

Unfortunately, only 2 papers (4\%) explicitly mentioned security measures during message communication. Some papers might have used the security features provided by the underlying platforms and operating systems used (e.g. ROS, RTOS, MRL), but this has not been specified by the authors explicitly. We argue that nowadays IoT offers a wide range of secure communication protocols \cite{nguyen2015survey} and they should be harnessed by researchers to increase the security of their robot implementations. IoT could also be used to enhance humans' access control via smart authentication, i.e. by means of various biometrics, face or voice recognition etc.

\subsection{Connection to the Internet/Web of Things}
\label{linkIoT}
Our research shows that some papers claimed to be IoT-enabled without giving any details of the connection between IoT and robotics \cite{kim2016self, mozaffari2016mobile, mozaffari2017mobile, ahn2018reliable}. Also, the majority of the surveyed papers (29 papers, 59\%) use merely TCP/IP communication, which is only a small part of the concepts of IoT and IoTR \cite{ray2016internet, vermesan2017internet, simoens2018internet}. In regards to the WoT, many of the surveyed papers (20 papers, 40\%) perform HTTP-based communication or provide only some basic HTTP interface for interacting with the robots.

The most complete papers in IoRT are those employing cloud robotics (see Table \ref{tab2}), while the most complete ones in WoRT are those using REST APIs for robot control and/or semantic web technologies (see Table \ref{tab3}). The observations that a limited number of papers employ IPv6 architectures or cloud robotics (12 papers, 24\%) and/or use REST APIs or semantics (5 papers, 10\%) are indications that the penetration of IoT/WoT in robotics is still low and that the IoT/WoT are still not largely and properly used in robotics. This phenomenon has been observed also in WoT frameworks in the past \cite{kamilaris2016web}, where the authors claimed to have WoT-ready frameworks, however their developments missed some important elements of the WoT principles.

As mentioned before, cloud networked robotics is a modern, promising research area  \cite{hu2012cloud}. New cloud-based software systems make the integration between robotics and IoT much easier \cite{hamblen2013embedded, toris2015robot}. The ROS could be used for connecting robots to the cloud \cite{quigley2009ros} (i.e. together with the FIROS tool \cite{limosani2019connecting}). Such possibilities are also provided by RoboEarth \cite{waibel2011roboearth}, a system for sharing knowledge between robots. Rapyuta, as the RoboEarth cloud engine, helps robots to offload heavy computation by providing secured customizable computing environments in the cloud \cite{hunziker2013rapyuta}. Towards WoRT, the work in \cite{keppmann2015building} allows to build REST APIs for ROS, while development of Web-based services for robotic devices is possible via \cite{mayer2012web}.

\subsection{The Big Picture}
\label{bigPicture}
To complete the discussion on the IoRT/WoRT topics, we try to capture here the big picture of this research area.

First, it is important to mention the performance measures that can be used to assess the performance and/or compare various robot systems \cite{steinfeld2006common}. Many performance metrics exist in order to assess the quality of a robot or a network/swarm of robots, but the ones most commonly adopted are power/energy consumption, bandwidth, latency, throughput, resilience to errors, packet loss and locality of program execution. Some metrics are more suitable for UAV \cite{ju2018multiple}, such as setup time, flight time, inaccuracy of land, haptic control effort and coverage ratio. Unfortunately, our analysis cannot get in depth on this aspect because none of the cited authors discussed, mentioned or performed evaluation of the proposed robotic systems using these assessment metrics and/or and comparison with existing similar robot implementations. The only exception was the work in \cite{scilimati2017industrial}, where the authors assessed the quality of service of their robotic system (i.e. UAV with Telosb sensors), expressed in terms of network joining time, data retrieval delay and packet loss ratio. Their conclusion was that the performance of the UAV based on the aforementioned metrics satisfied the mission requirements.

Generally, authors of work under study preferred to focus their research on the feasibility and demonstration of connecting robots to the Internet/Web, as well as to the new possibilities (i.e. sensing, actions, application domains) that arose by their robot models/implementations (see Figure \ref{fig1}). We acknowledge the fact that in many cases there had not been any similar robotic systems to compare with.

It is still important to quantify and assess the impact of the TCP/IP stack and/or Web technologies to the real-time operations of robotic systems. The papers under study have not discussed this issue. Although other research works demonstrated that the impact of IoT/WoT on embedded devices is low \cite{kamilaris2011homeweb, jara2012glowbal, schor2009towards, dunkels2009efficient, rodrigues2010survey}, this might not be the case for time-critical robots.

\begin{figure*}
  \centering
  \includegraphics[width=1.0\linewidth]{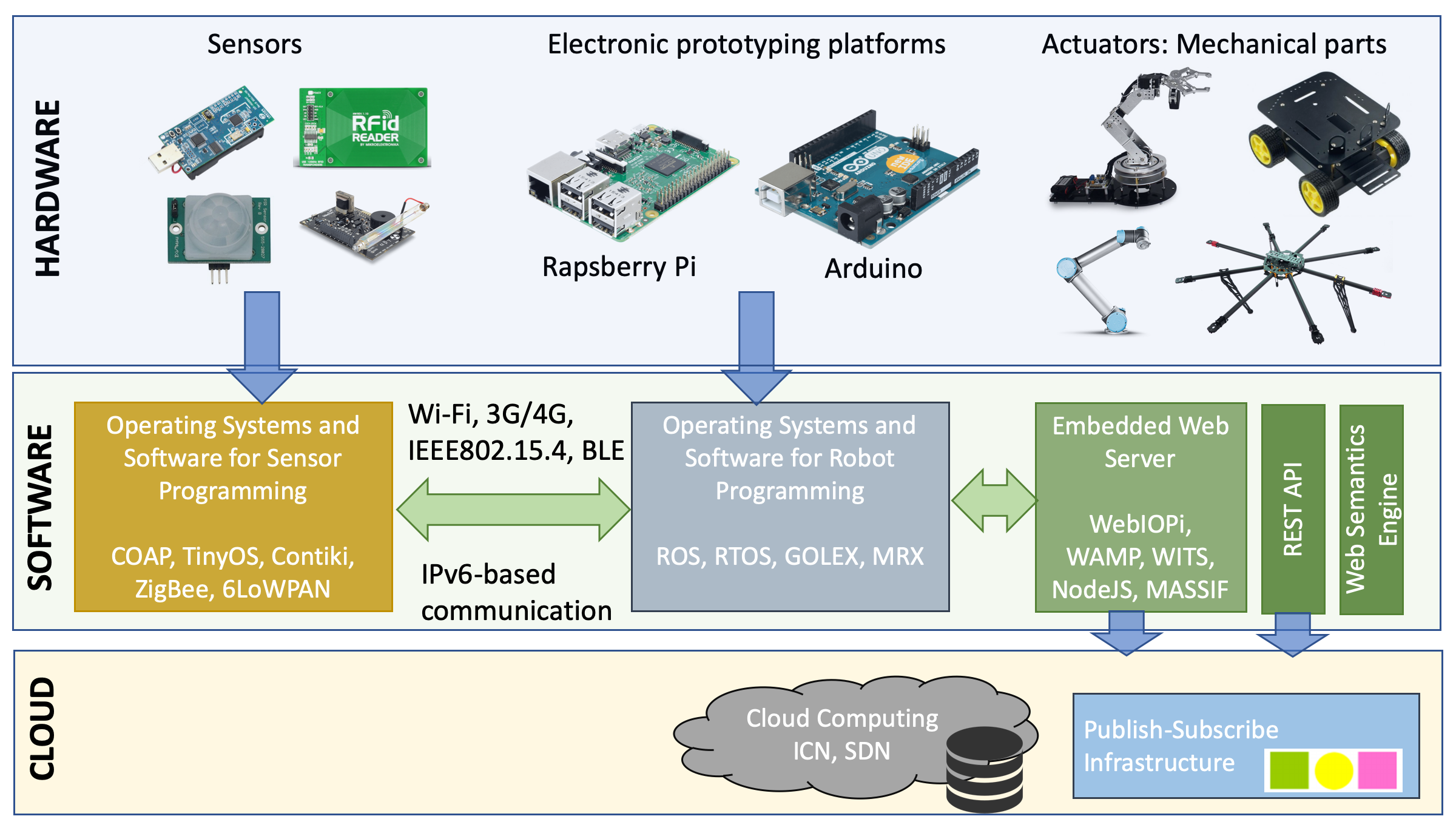}
\caption{The general architecture for IoRT/WoRT systems.}
\label{fig:IoRTarch}
\vspace{-0.5cm}
\end{figure*}

Figure \ref{fig:IoRTarch} depicts the general suggested architecture for IoRT/WoRT systems, considering the principles of IoT and WoT applied in robotics. Communication between sensors and electronic platforms is via the IPv6 protocol, while it is possible for sensory platforms to embed Web servers themselves \cite{kamilaris2011homeweb}. Figure \ref{fig:IoRTarch} lists some of the operating systems, as identified in related work under study for sensors' and robots' programming, or for deploying Web servers on the robots.

Some promising platforms for robots' programming not employed in the surveyed papers but still worth mentioning are FIROS \cite{limosani2019connecting}, a tool for connecting mobile robots to the cloud (i.e. by using ROS), BrainOS\footnote{BrainCorp, BrainOS. https://www.braincorp.com/brainos-autonomous-navigation-platform}, an autonomous navigation platform based on computer vision and artificial intelligence, as well as the Middleware for Robotic Applications (MIRA) \cite{einhorn2012mira}, a cross-platform framework for the  development of robotic applications.

Towards full Web integration, robots may expose their services as a REST API, for easier reuse of their capabilities and features by authorized third parties. Robots may also incorporate semantic engines for annotating semantically their services and data, towards seamless M2M interaction. As mentioned in Section \ref{webSemant}, researchers employed various ontologies to describe robotic data (i.e. WSN, SSN) and robotic services (i.e. Brizzi et al. \cite{brizzi2013bringing}). A set of interesting ontologies that could be used in future IoRT systems is the IEEE Ontologies for Robotics and Automation (ORA) \cite{prestes2013towards}. These ontologies could also describe the relationship between a robot agent and its physical environment. As an example, Jorge et al. \cite{jorge2015exploring} used the ORA ontology for spatial reasoning between two robots that must coordinate for providing a missing tool to a human. Another interesting set of semantics is proposed in \cite{tenorth2012unified}, for projecting the effects of actions and processes performed by the robots and their sequence. Besides ontologies, query languages and engines (e.g. SPARQL, CQEL) are important for semantic reasoning.

For completing IoT integration, especially in cases where robot swarms are involved or high scalability/performance are needed, cloud services could be the solution. The cloud could be used for more advanced processing and for storage of information, but also for efficient messaging via publish/subscribe infrastructures. Concepts such as ICN \cite{ahlgren2012survey} and SDN  \cite{mckeown2009software} could also be employed for better overall management and control of the robots in a more abstract and generic manner.

As mentioned in Section \ref{meth}, comparison of IoT/WoT existing software and hardware platforms is out of the scope of this paper. It is worth commenting however that the inclusion of IoT/WoT protocols gives important benefits to (not only) robotic systems, such as interoperability, seamless M2M communication, easier integration to existing systems and infrastructures, as well as use of well-known and popular technologies for programming, management and control \cite{wilde2007putting, guinard2011internet}. Some of these benefits have been highlighted in Sections \ref{iotrobbenefits} and \ref{woTRbenefits}. As mentioned before however, more studies are needed in order to further validate the impact of the IoT/WoT protocols and technologies on the real-time performance of robots.

\subsection{Challenges and Future Opportunities}
\label{future}
Robotics, together with IoT, constitute dynamic and active research fields and there is much ongoing research in these areas. This section focuses on the existing challenges and barriers, as well as future research opportunities that arise from the combination of these technologies together in the future.
Some of these challenges and/or opportunities are the following:
\begin{itemize}

\item Although the impact of IoT/WoT technologies and protocols on embedded devices may be low \cite{kamilaris2011homeweb, jara2012glowbal, schor2009towards, dunkels2009efficient, rodrigues2010survey}, this impact might still be considerable in time-critical robotic systems, where decisions need to be taken in fragments of a second. We welcome research efforts dealing with this aspect, studying in detail how different technologies and protocols of IoT and WoT affect performance. This could be well related to the security of the messages exchanged between robots and infrastructures \cite{suo2012security}.

\item As mentioned in Section \ref{bigPicture}, it is unfortunate that no comparisons of the robots with existing literature in terms of performance metrics was performed in the related work under study. We expect that future research will focus more on performance, studying the possibilities and constraints of the robots considering low-level metrics such as energy consumption and autonomy, bandwidth, latency, throughput, resilience and fault tolerance, as well as high-level metrics such as ease of use and control of the robot, overall engagement and social behaviour, safety etc. It would be also important to study the penalty and trade-offs in energy consumption and autonomy when Web servers and the IPv6 stack are added to the robots \cite{jara2012glowbal, kamilaris2011homeweb}.

\item In relation to the above, an important challenge( and desired characteristic) of robots is autonomy in operation. IoT has active research in breakthrough high-performance architectures, algorithms and hardware that will allow wireless networks to be highly efficient \cite{mahmoud2016study}, powered by tiny batteries, energy-harvesting \cite{kamalinejad2015wireless}, or over-the-air power transfer. Moreover, new communication systems based on biology and chemistry are expected to evolve in IoT research, enabling a wide range of new micro- and macro-scale applications \cite{akyildiz2010internet}.

\item Another important research challenge for robots is the collective behaviour of robot swarms, or the coordination and control of multi-robot systems towards an optimized outcome. Such swarms have been proposed for collective industrial construction \cite{petersen2011termes}, transportation and box pushing \cite{parker2006building},
as well as for smart homes \cite{saffiotti2005peis} and search and rescue \cite{das2007mobility}. More generally, the authors in \cite{lundh2007dynamic} propose a plan-based approach, based on PEIS Ecology \cite{saffiotti2005peis}, to automatically generate a preferred configuration of a robot ecology given a task, environment and set of resources. In this robot swarm ecosystem, IoT can offer effective solutions for networking, communication among the robots and mobility \cite{das2007mobility}.

\item Just like there exist complete operating systems (e.g. TinyOS, Contiki) and application-layer protocols (e.g. COAP, ZigBee) for IoT sensors and devices, we expect similar operating systems to appear for robots \cite{pwcReport}. As mentioned before, early efforts in this direction are BrainOS and MIRA. A culture of developers and researchers still needs to be built around these platforms/frameworks, sharing code, experiences and solutions to the community. Worth mentioning is the \textit{Brains For Bots} SDK by Neurala\footnote{Neurala, Brains For Bots. https://www.welcome.ai/tech/hardware-iot/neurala-brains-for-bots-sdk}, which includes numerous features for creating applications that can learn, recognize, find and track objects in real-time. Neurala incorporates in its SDK deep learning techniques, to support cognitive requirements in robot applications.

\item Combining online social networking together with IoT/WoT-enabled robots \cite{turcu2012social}. In this case, online social networks could be used for storing and sharing links to resources of interest for the R2R and R2H interactions, facilitating sharing of robotic services among online trustful contacts \cite{kamilaris2012practice}. Robots could recognize or authenticate users, giving them access to some of their controls, based on the peoples' online profiles and endorsements they have from other authorized people.

\item In industrial applications and logistics, technologies that blend the physical and digital context are important. Broader and more expansive information capture and processing via IoT/WoT, combined with smarter manipulation and movement of physical materials via robots, can deliver new benefits such as higher efficiency of operations, more insights and visibility, as well as better interaction between components, systems and actors. Blockchain could be relevant, to keep an immutable distributed ledger among IoT sensors and robotic systems of untrusted partners involved in some supply chain ecosystem \cite{tian2017supply, vermesan2017internet}.

\item The trend towards human-centric design of robotic systems is expected to continue \cite{pwcReport} and robots will become more integrated in our everyday lives, either for assistance in common tasks or for actual support of people in need, e.g. movement of people with paralysis. Thus, they need to capture humans' emotions and social behaviour to understand how to react in different situations \cite{fong2003survey}. This information might come from IoT sensors, either installed on the robots or in the nearby environment (i.e. smart homes/buildings) or from the online social networking presence and activity of humans. Ethics is an important dimension in this direction \cite{kortner2016ethical} and needs to be considered a priori. 

\item Towards human-centric robotics, the teaching of robots to take actions by means of natural language instructions is a key aspect. This challenge could derive knowledge and ideas from Web Semantic technologies, as well as from ontology-based natural language interfaces for controlling IoT devices, such as im4Things \cite{noguera2017im4things} and home automation \cite{baby2017home}.

\item We expect service robots to enter new real-world environments in which IoT has already penetrated \cite{KamilarisIeeeIoT16}. These environments could be sports (i.e. assistance and safety in sports having risks such as climbing and parachuting), health (i.e. not only monitoring patients but also taking first-aid actions if needed), ecology and environmental monitoring, surveillance and security in social spaces (i.e. schools, airports, urban hotspots etc.), gaming (i.e. robots become comrades or opponents in real-world gaming scenarios), customer service (i.e. restaurants, hotels, gyms, tourism), the movie industry and many others. The autonomous learning capabilities of  robots, together with their potential actions due to their mechanical parts, could offer more opportunities in these new environments.

\end{itemize}
 

\subsection{Demonstration of an IoT Robot for Pipeline Inspections}
\label{DemoPIRATE}
A relevant application domain where robotics would be very useful is industrial inspection and maintenance, using autonomous robots. In the context of petrochemical pipeline plants, in-pipe inspection can be performed using snake-like robots \cite{dertien2014design}.
 
\begin{figure*}[h]
   \centering
   \includegraphics[width=0.90\linewidth]{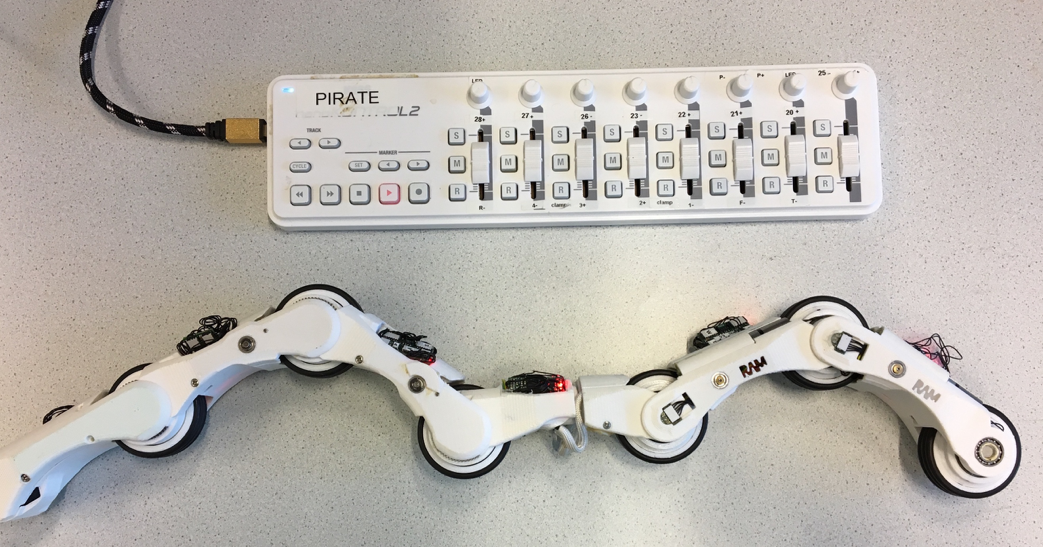}
   \caption{The PIRATE robot: It has a multi-links body in which each segment is coupled to the next one using an actuated rotational joint.}
   \label{fig:pirate1}
\end{figure*}

The first steps in the direction of autonomous navigation and inspection have been taken at the Robotics and Mechatronics Department of the University of Twente. This is a new application in IoT-related robotics, according to Table \ref{tab1}. The process of crawling is also a new action according to Figure \ref{fig1}, based on the capability of moving.
Figure \ref{fig:pirate1} presents the PIRATE (Pipeline Inspection Robot for AuTonomous Exploration) robot, which is designed to travel through pipes having different diameters, vertical sections and sharp corners. The robot is composed of six modules connected through actuated rotational joints. Each module also includes a driven wheel. 

\subsubsection{Hardware}
\begin{figure*}[h]
   \centering
   \includegraphics[width=0.60\linewidth]{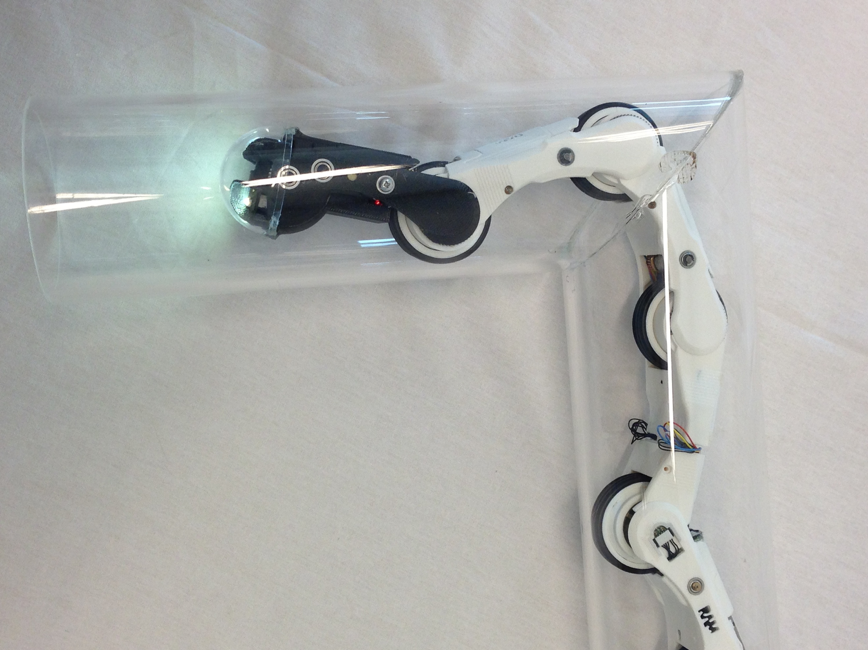}
   \caption{The PIRATE robot taking a sharp 90 degrees turn.}
   \label{fig:pirate2}
\end{figure*}

PIRATE is equipped with absolute encoders for measuring the joints angles and consequently infer its configuration in the pipe and the wheels' speed, with inertial measurements units (IMU) for measuring the acceleration and orientation. Thanks to the encoders and the IMU, it is possible to infer the pose (i.e. position, joint configurations and orientation of the robot) in the pipe. This info is necessary for the autonomous control of the robot, especially for difficult moves such as taking a sharp 90 degrees turn, as shown in Figure \ref{fig:pirate2}.

Moreover, the robot is enhanced with cameras for visual inspection, with Light Detection and Ranging (LiDAR) sensors for navigation and ultrasonic sensors for measuring the wall thickness. The latter is an important parameter for assessing the state of the pipe. LiDAR and ultrasonic constitute new sensing equipment for IoT robots, according to Figure \ref{fig1}.

\begin{figure*}[h]
   \centering
   \includegraphics[width=0.90\linewidth]{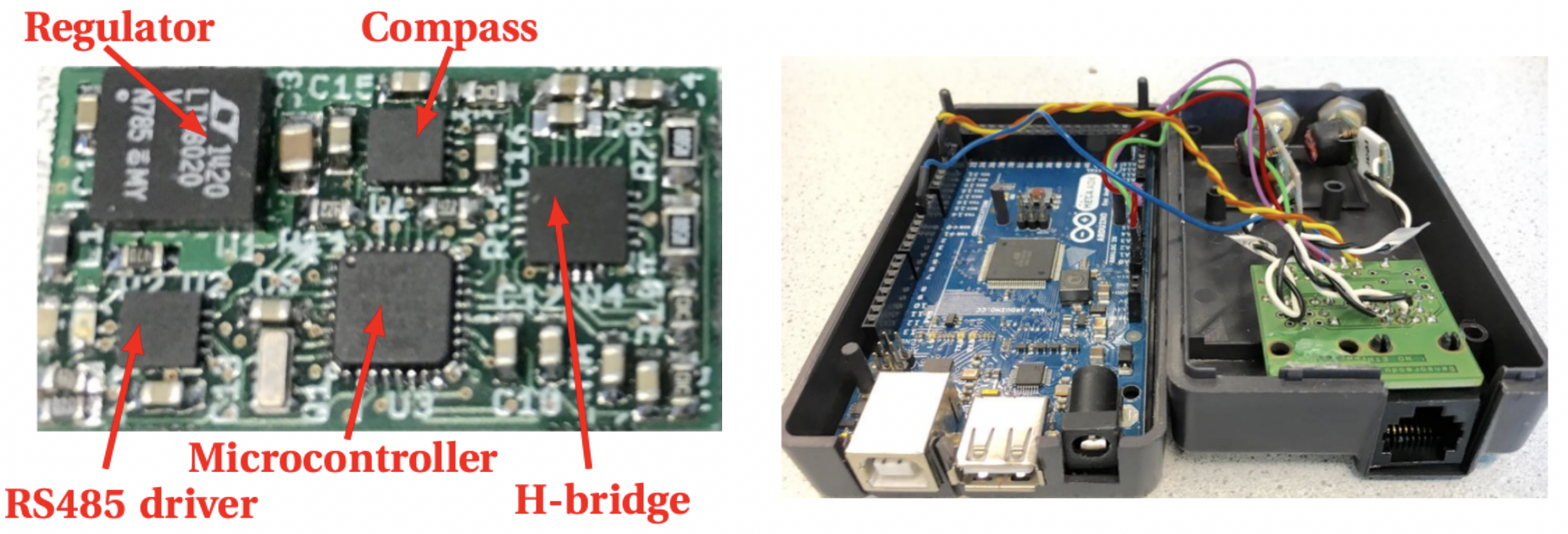}
   \caption{The control board (on the left) and the Arduino board (on the right) as used on the PIRATE robot.}
   \label{fig:pirate3}
\end{figure*}

The robot is controlled via an Arduino board that is used as transparent bridge between the laptop which is used for programming and the robot and the (custom) boards on the robot. These boards directly control motors and read the information from the sensors. On each board, a microcontroller, a current regulator, an H-bridge, a RS485 driver and an IMU compass are installed. This configuration is depicted in Figure \ref{fig:pirate3}.

The robot is currently wired (via an Ethernet cable), due to the fact that that popular wireless protocols did not function well inside pipe systems (e.g. IEEE802.15.3, Wi-Fi, BLE), according to our observations. Theoretically, \textit{super low frequency carrier} wireless communication modules can be used \cite{zangl2008investigation}, but we did not possess this equipment. It is an interesting aspect of future work.

\subsubsection{Software}
The software code has been developed on top of ROS \cite{quigley2009ros}, based on C++. ROS was selected because it offers great flexibility for designing software architectures, offering many libraries and algorithms for tasks such as motion, object detection, mapping, etc. (see Section \ref{software}). Its main disadvantage is that it does not support real-time guarantees, but this was not a requirement of our deployment.

An important characteristic of ROS is that it offers, thanks to FIROS \cite{limosani2019connecting}, the possibility to connect robots and robotic services to the cloud. FIROS provides a publish-subscribe scheme and a REST API for managing robot connections and to receive the data published by the robot (i.e. publisher) via our laptop (i.e. subscriber). 

Specifically, we used the following steps to enable PIRATE to the Internet/Web via FIROS:
\begin{enumerate}
 \item First, we added the PIRATE robot to the whitelist of FIROS (i.e. a whitelist.json file) via the \textit{POST /whitelist/write} API call, in order to make PIRATE discoverable and manageable.
 \item Then, we used the \textit{POST /robot/connect} API call to connect to the PIRATE robot. This call makes FIROS to connect to to new robots in case their names and topics match the ones allowed on the whitelist.json file.
 \item Finally, the \textit{GET /robot/PIRATE} API call was used to receive the robot's sensory information in a JSON format.
\end{enumerate}

The performance of the robot in terms of task efficiency, reliability, sensor accuracy, response time and energy consumption constitutes subject of future work and it will be topic of another publication.

\subsubsection{Potential Benefits}
In addition to the general benefits of enabling robots to the IoT (see Section \ref{iotrobbenefits}) and to the WoT (see Section \ref{woTRbenefits}), in the context of the PIRATE robot this IoT/WoT enabling allowed the constant and continuous interaction and communication between the robot and human operators via a well-known and understood medium, i.e. the Internet. During the inspection mission, the robot gathered data about the environment (i.e. the pipe), which were described using semantic Web technologies via the SSN ontology \cite{compton2012ssn}. Using this ontology (or other relevant ones), researchers may understand and reuse data collected by PIRATE without much effort and ambiguity. Using the Internet/Web as medium for gathering and storing data allows also to process and examine this data remotely and in real-time. Operators who may be located in different regions of the world may work together to indicate different inspection targets to PIRATE. Real-time Internet connectivity and data acquisition facilitates even the deployment of swarms of in-pipe robots, which may be deployed to increase speed and efficiency of the inspection procedure, while the communication provided by IoT protocols is crucial for the optimization of the task. The latter constitutes current work of the Robotics and Mechatronics Dept., aiming to develop IoT-enabled robotic swarms which allow real-time monitoring and control via the Internet/Web.

\subsection{Take-Home Messages}
\label{takeAwayMessages}
Summarizing the study performed in this paper, some take-home messages can be the following:
\begin{itemize}
 \item As the IoT penetrates different domains, application areas and scientific disciplines, it started to penetrate also the research area of robotic systems.
  \item There is a wide range of sensors used, intended actions of the robots and application areas where the IoRT/WoRT-based robots operate. The most popular actions involve moving, observing via computer vision, flying and navigating. The most popular application areas are entertainment, health, education and surveillance.
  \item A large percentage of the surveyed papers employed open-source hardware and prototyping platforms, making the physical connection to a wide variety of different sensors possible.
  \item Regarding communications, various protocols are being used, with Wi-Fi being the most popular, followed by IEEE802.15.4 and Bluetooth. The decision on which communication protocol to use was mainly influenced by the coverage range required, as well as the need for autonomy and longer lifetime of operation (i.e. energy consumption).
 \item The IoT is still not used at its full potential in robotics. Cloud robotics and the IPv6 protocol are not fully utilized. The same holds for the principles of WoT, which has not yet been fully exploited in robotic systems. Services of the robots are not generally exposed as a REST API, nor the services and data are described via Web semantic technologies.
 \item There is an increasing number of operating systems, software tools and platforms for developing IoRT/WoRT-based robotic systems. Most of this software enables access to cloud computing and publish-subscribe infrastructures for scaling computation and storage capabilities and for supporting robot swarms more easily.
  \item Related work mostly focuses on feasibility studies and demonstrations, which showcase the benefits of IoT/WoT in robotics, mainly in terms of ubiquitous access via Internet/Web and easy interoperability with other systems. Aspects of performance, especially considering the overhead produced by the TCP/IP protocol, have not been well studied.
 \item Although security and privacy constitute important aspects of robots and their services/data, they have not been addressed by the works under study.
 \item There are still numerous open issues and gaps for future research in this emerging intersection of research areas. Important research directions relate to the presence of robots in online social networks, autonomy in operation, collective behaviour of robot swarms, taking actions by means of natural language, deriving knowledge from Web semantics, design of more empathic, social, human-centred service-based robots and others. 

\end{itemize}

\section{Conclusion}
\label{conclusion}
This paper studied the current use of the Internet of Things (IoT) in robotics, through various real-world examples encountered through a research based on a bibliographic-based method. The concepts, characteristics and architectures of IoT, as they are being used in existing robotic systems, have been recorded and listed, together with popular software, hardware and communication methods. Moreover, the application areas, sensors and robot services/actions incorporated in IoT-based robots are presented. Further, the current application of the Web of Things (WoT) in robotics has been investigated and the overall potential of the Web of Robotic Things has been discussed in the paper. A general observation is that some of the advanced concepts of IoT/WoT are not yet being used by researchers in robotics. Finally, future research directions and opportunities are proposed.

\section*{Acknowledgements}
Andreas Kamilaris has received funding from the European Union's Horizon 2020 research and innovation programme under grant agreement No 739578 complemented by the Government of the Republic of Cyprus through the Directorate General for European Programmes, Coordination and Development.

Nicol\`{o} Botteghi has received funding from Smart Tooling. Smart Tooling is an Interreg Flanders-Netherlands project sponsored by the European Union focused on automation in the process industry: making maintenance safer, cheaper, cleaner, and more efficient by developing new robot prototypes and tools.

\bibliographystyle{plainnat}
\bibliography{robob}   

\end{document}